\documentclass[10pt,twocolumn,letterpaper]{article}

\usepackage{cvpr}              

\usepackage[dvipsnames]{xcolor}

\usepackage{url}
\usepackage{multirow}
\usepackage{nicematrix}
\usepackage{tabularx}
\usepackage{xcolor}
\usepackage{soul}
\definecolor{cvprblue}{rgb}{0.21,0.49,0.74}
\usepackage[pagebackref,breaklinks,colorlinks,citecolor=cvprblue]{hyperref}

\def\paperID{2} 
\def\confName{CVPR}
\def\confYear{2024}

\title{CattleFace-RGBT: RGB-T Cattle Facial Landmark Benchmark}

\author{Ethan Coffman$^{\ast}$, Reagan Clark$^{\ast}$, Nhat-Tan Bui$^{\ast}$, Trong Thang Pham$^{\ast}$, \\ 
Beth Kegley$^{\dagger}$, Jeremy G. Powell$^{\dagger}$,Jiangchao Zhao$^{\dagger}$, and Ngan Le$^{\ast}$ \\
$^{\ast}$AICV Lab, University of Arkansas, Fayetteville, Arkansas, USA \\
$^{\dagger}$Department of Animal Science, University of Arkansas, Fayetteville, Arkansas, USA \\
}
\begin{document}
\maketitle

\begin{abstract}

Ensuring the well-being of cattle is of utmost importance for both ethical and economic reasons. An intelligent system that can accurately assess the physical state of cattle would provide invaluable insights for farm management to optimize animal welfare.
Cattle express their well-being through various signals, e.g. body temperature and facial expression across different facial parts such as watering eyes, runny nose, etc. However, these indicators has traditionally relied on human observation with manual measuring which is time-consuming and subjective.

To address this challenge, we introduce CattleFace-RGBT, a RGB-T Cattle Facial Landmark dataset consisting of 2,300 RGB-T image pairs, a total of 4,600 images. 
Creating a landmark dataset is time-consuming, but AI-assisted annotation can help. However, applying AI to thermal images is challenging due to suboptimal results from direct thermal training and infeasible RGB-thermal alignment due to different camera views. Therefore, we opt to transfer models trained on RGB to thermal images and refine them using our AI-assisted annotation tool following a semi-automatic annotation approach. 
Accurately localizing facial key points on both RGB and thermal images enables us to not only discern the cattle's respiratory signs but also measure temperatures to assess the animal's thermal state. To the best of our knowledge, this is the first dataset for the cattle facial landmark on RGB-T images. We conduct benchmarking of the CattleFace-RGBT dataset across various backbone architectures, with the objective of establishing baselines for future research, analysis, and comparison. The dataset and models are at \href{https://github.com/UARK-AICV/CattleFace-RGBT-benchmark}{https://github.com/UARK-AICV/CattleFace-RGBT-benchmark}
\end{abstract}    
\section{Introduction}
\label{sec:intro}



In recent years, the AI research community has shown a growing interest in measuring and understanding animal expressions and welfare alongside those exhibited within the human domain through various studies~\cite{liu2012dog,jinkun19cross,animalweb,catflw}. Despite notable advancements, cattle datasets are limited, which is an important factor related to the development of livestock farming and the agricultural industry. Although CattleEyeView \cite{cattleeye} provides images about cattle, it has limited information. The top-down view multi-task cattle video dataset offers restricted information about signs of poor health, such as tiredness or excessive salivation. Moreover, it only includes RGB frames, which cannot provide temperature data. 
Thermal images are essential for identifying cattle sickness, as many illnesses are associated with changes in body temperature. The lack of thermal data in existing datasets poses challenges for effectively monitoring and assessing the physical health of cattle, particularly in detecting fever.

\begin{figure}[t]
\centering
\setkeys{Gin}{width=\linewidth}
\resizebox{\linewidth}{!}{%
\begin{tabular}{cc}
     \includegraphics{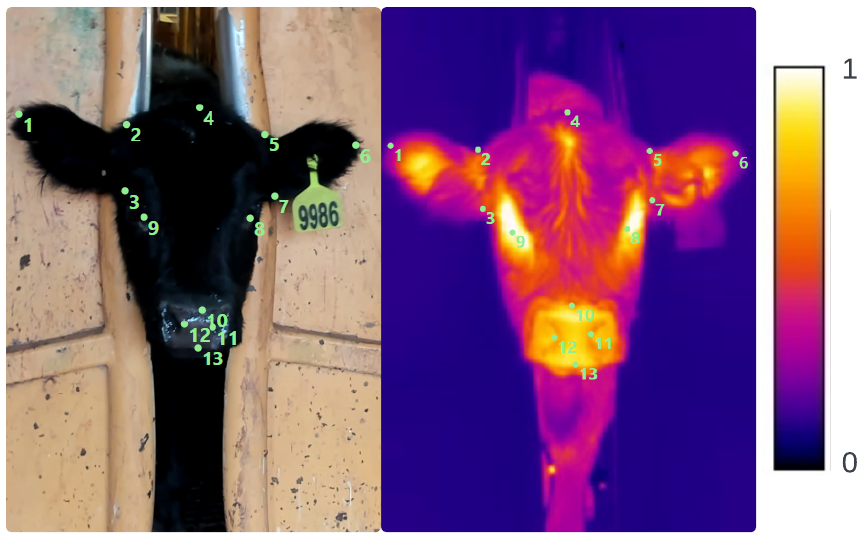} \\
\end{tabular}}
\caption{Example of RGB and thermal pair image with 13 annotated keypoints annotated in our CattleFace-RGBT dataset.}
\label{fig:example_cattle}
\vspace{-5mm}
\end{figure}

\begin{table*}[!h]
\centering
\setlength{\tabcolsep}{5pt}
\renewcommand{\arraystretch}{1.0}
\resizebox{\linewidth}{!}{
  \begin{tabular}{@{}l|llcllll@{}}
    \toprule
    \textbf{Datasets} & \textbf{Animal} & \textbf{Images} & \textbf{\# Keypoints} & \textbf{Data Modalities}  &   \textbf{Focus} & \textbf{Objectives} \\
    \midrule
    Animal Pose \cite{jinkun19cross} & Various & 4,000+ & 20 & RGB &  Body-Head & Pose estimation \\
    AnimalWeb \cite{animalweb} & Various & 22,451 & 9 & RGB & Head & Facial behaviour monitoring\\
    Zhang et al. \cite{cathead} & Cat & 10,000 & 9 & RGB & Head & Head detection\\
    CatFLW \cite{catflw} & Cat & 2,016 & 48 & RGB &  Head & Facial landmark detection\\
    Liu et al. \cite{liu2012dog} & Dog & 8,351 & 8 & RGB  & Head & Breed identification\\
    Yang et al. \cite{yang15sheep} & Sheep & 600 & 8 & RGB &  Head & Facial landmark detection\\
    Horse Facial Keypoint \cite{horse} & Horse & 3,717 & 5 & RGB & Head & Facial landmark detection\\ 
    CattleEyeView \cite{cattleeye} & Cattle & 30,703 & 24 & RGB &  Body & Various\\
    \midrule
   \textbf{CattleFace-RGBT (ours)} & Cattle & 4,600 & 13 & RGB \& \textbf{Thermal} &  Head & Facial landmark detection\\
   \bottomrule
  \end{tabular}}
  \caption{Data properties comparison between our work with previous animal landmark datasets. }
  \label{tab:dataset_compare}
\end{table*}

To overcome the limitations of existing datasets, this paper introduces the CattleFace-RGBT dataset, which is the first of its kind to include both RGB and thermal cattle images captured from a frontal view, along with 13 keypoint landmarks. 
Our dataset, collected directly from a farm, is annotated using an AI-assisted method and manually corrected to ensure accuracy. Given the time-consuming and elaborate nature of creating a landmark dataset, utilizing AI-assisted annotation will greatly reduce the time and effort involved. It is important to note that there are two main challenges when applying an AI model to annotate thermal images. First, training models directly on thermal often yield suboptimal results~\cite{li2023feasibility}. 
Second, perfect automatic alignment between RGB and thermal is infeasible due to different camera views, so we cannot directly transfer keypoints from RGB to thermal images. To address the challenges, we leverage the knowledge gained from the RGB domain and adapt it to the thermal domain so that we can accelerate the annotation process while maintaining high accuracy, despite the challenges posed by thermal imaging data and different camera views. More details are described in \cref{sec:ai-assisted-annotation}.



The comparison between our CattleFace-RGBT with the existing animal datasets is given in Table \ref{tab:dataset_compare}. It can be seen that the CattleFace-RGBT dataset is the first dataset for cattle facial landmark detection on thermal images. Figure \ref{fig:example_cattle} shows an example of the pair RGB and thermal cattle images in our dataset. For every pair in the dataset, the cattle is annotated with 13 keypoints on its ears, poll, eyes, muzzle, nostril, and mouth (\cref{sec:dataset}).
Finally, we conduct experiments on various state-of-the-art keypoint detection models to assess the effectiveness of existing approaches and provide a baseline for future work in cattle facial landmark detection and analysis. 

In summary, our contribution is two-fold:
\begin{itemize}
    \item To the best of our knowledge, we introduce the first dataset containing both RGB and thermal images for cattle. It includes 2,300 RGB and thermal pairs (4,600 in total). We provide 13 key points in both RGB and thermal images on key cattle facial parts, i.e. eyes, ears, muzzle, nostril, and mouth. By releasing this novel dataset, we aim to advance research in the field of cattle welfare.
    \item We benchmark the dataset on various backbone networks to demonstrate the performance of existing methods to estiblish baselines for future research, analysis, and comparison.
\end{itemize}

\section{Related Datasets}
Although the field of animal landmark has not been well studied like the human domain \cite{6755925,6130513,deng2019menpo}, various datasets have been proposed. Animal Pose \cite{jinkun19cross}, that contains 4000+ images with 20 keypoints of 5 species, is extended based on the subset of VOC2011 \cite{pascal-voc-2011}. AnimalWeb \cite{animalweb} consists of 22,451 images of 350 different animal species faces with 9 fiducial landmarks. Zhang et al \cite{cathead} pays attention to the cat-like animal when introducing 10,000 images with 9 facial landmarks. After that, CatFLW \cite{catflw} uses the subset of original Zhang et al.~\cite{cathead} dataset to increase the number of landmarks, which introduce the dataset contain 2016 images with 48 landmark points. Liu et al. \cite{liu2012dog} focuses on dog breeds by introducing the dataset contains 8,351 images of 133 dog breeds. In addition, Yang et al. \cite{yang15sheep} and Horse Facial Keypoint \cite{horse} report the datasets of 600 sheep images and 3717 horse images, respectively.
Among datasets collected for animal species, we are mostly related to CattleEyeView \cite{cattleeye}, where the authors provide the top-down view of 30,703 cattle images. In essence, it explores the different applications (i.e., tracking and counting) of cattle images than ours. To the best of our knowledge, we are the first to propose the cattle dataset contains both RGB and thermal images in the front view. With 6,400 total images of 2,300 RGB and 2,300 thermal together with 13 keypoints, our dataset has the potentials for understanding the cattle welfare and detecting cattle fever, which fulfill the needs for smart farming.
\section{The CattleFace-RGBT Dataset} 
\label{sec:dataset}

\begin{figure}[t]
\centering
     \includegraphics[width=\linewidth]{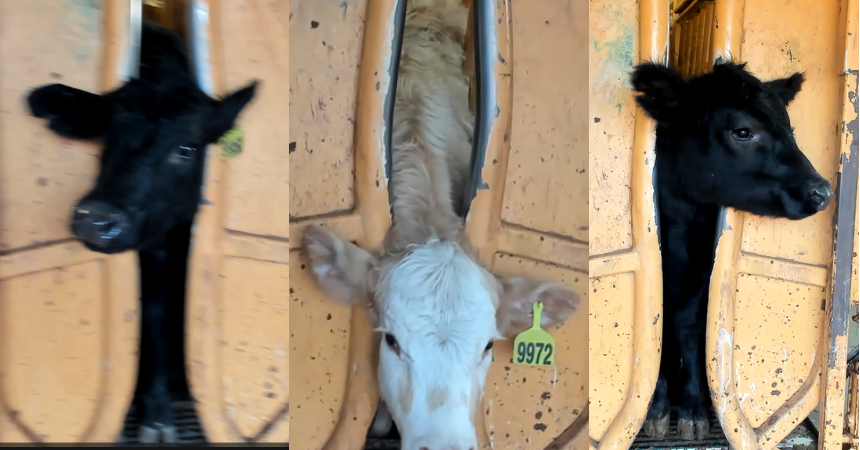}
\caption{Examples of undesired images.}
\label{fig:undesired_image}
\end{figure}

\begin{figure}[t]
\centering
\includegraphics[width=\linewidth]{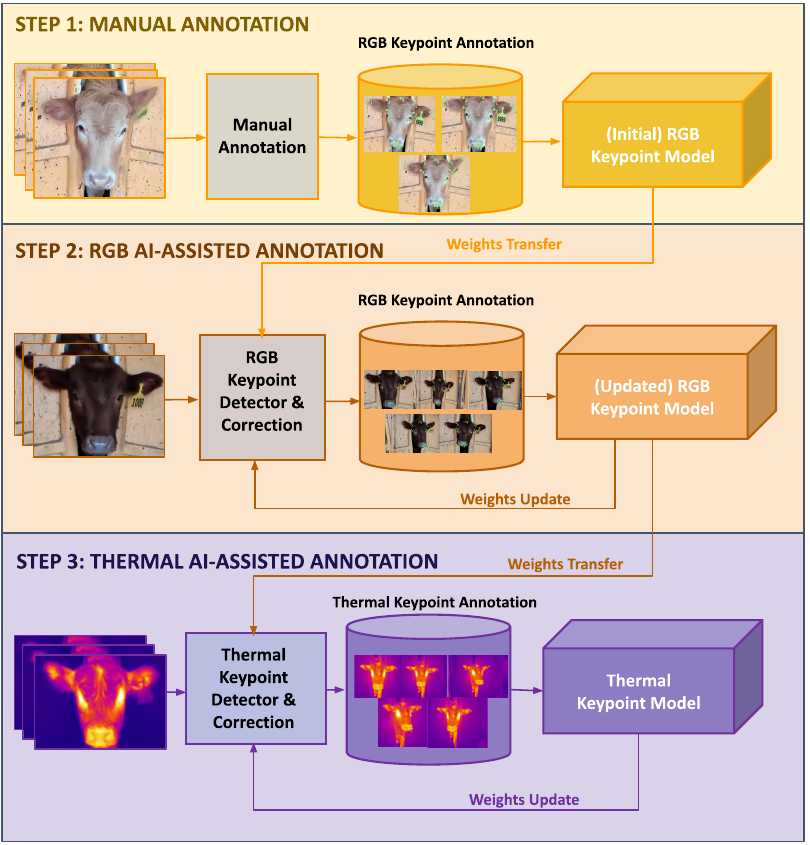}
\caption{The workflow for creating our dataset.}
\label{fig:process}
\end{figure}
The CattleFace-RGBT Dataset is a curated selection of videos captured in collaboration with agricultural researchers. Before annotation, undesired images were filtered from the raw data, as described in \cref{sec:preprocess_data}. The data is then annotated using a semi-automated approach, which involved manual annotation using a custom annotation tool and AI-assisted annotation, as detailed in \cref{sec:ai-assisted-annotation}. The landmark convention and annotation tool are described in \cref{sec:annotation_tool} and \cref{sec:landmark}, respectively.

\subsection{Image Collection and Preprocessing}
\label{sec:preprocess_data}
During the facility's standard calf processing, provided by agricultural researchers, we obtain thermal and RGB video footage. While each calf is temporarily restrained in a cattle squeeze chute, we record approximately 20 seconds of footage per calf using a tripod-mounted thermal camera and webcam, and concurrently acquire data on each calf's health status from the facility's records.


The final step before annotation involves filtering out undesired images. Since the primary objective of our dataset is to monitor animal welfare, we only retain images depicting cattle in a clear frontal view. We eliminate all frames in which the calf is not fully visible within the image boundaries.
This selective approach ensures that the dataset is tailored to our specific purpose and eliminates irrelevant images. Examples of undesired images are illustrated in Figure \ref{fig:undesired_image}.

\subsection{AI-assisted Annotation} 
\label{sec:ai-assisted-annotation}
As shown in \cref{fig:process}, our AI-assisted annotation pipeline involves a three-stage process: first, we perform manual annotation; second, we apply AI assistance on RGB images; and finally, we leverage AI assistance on thermal images.

\noindent
\textbf{Manual Annotation.} Our dataset consists of a total of 2,300 RGB and thermal image pairs, which we divide into four batches, each containing approximately 600 image pairs. We begin by annotating the RGB images first. In particular, 600 RGB images are manually labeled to form the first batch. After the manual annotation of the first batch, the images are fed into a deep-learning-based landmark detection model, specifically a Feature Pyramic Network (FPN)~\cite{fpn} with a ResNet50 backbone.

\noindent
\textbf{RGB AI-assisted Annotation.}
Once the model has been trained on the first batch in the previous step, the checkpoint from the first step is used for inference on the second batch. The model's predictions for the second batch are then manually corrected. Then, we use this corrected second batch to finetune our current keypoint detection model. Finally, we use this model to infer keypoints on the third batch. This loop continues for the subsequent batches, with the FPN model being fine-tuned using the corrected annotations from the previous batch. 

\noindent
\textbf{Thermal AI-assisted annotation.} The final checkpoint from the entire RGB annotation process is used as a pretrained model to predict keypoints on thermal images, eliminating the need for manual annotation of the first 600 thermal images. We then perform the same loop as with the RGB images until the entire dataset is annotated.

By employing this semi-automatic approach, we create a comprehensive facial landmark dataset while minimizing the manual annotation effort and ensuring high-quality annotations through expert correction.


\begin{figure}[t]
\centering
\includegraphics[width=\linewidth]{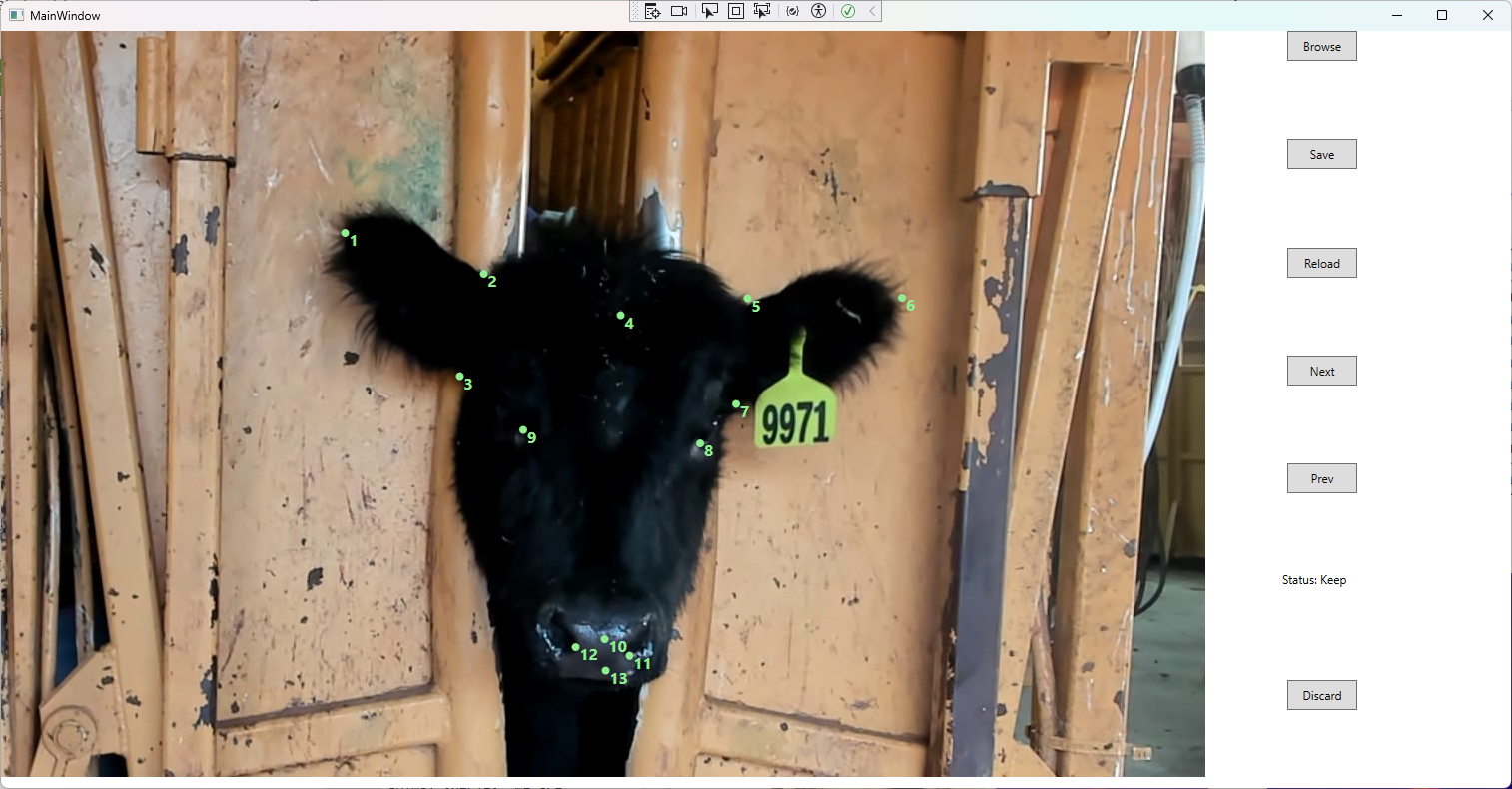}
\caption{The user interface of our annotation tool with the utilities on the right including ``Browse'',``Save'', ``Reload'', ``Next'', ``Prev'', ``Discard''.}
\label{fig:tool}
\end{figure}

\begin{figure}[t]
\centering
\includegraphics[width=\linewidth]{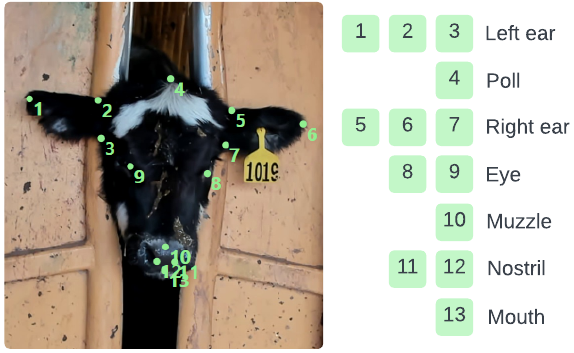}
\caption{Illustration of 13 keypoint landmarks.}
\label{fig:face_part}
\end{figure}

\begin{table}[t]
\setlength{\tabcolsep}{5pt}
\renewcommand{\arraystretch}{1.0}
\centering
\caption{Benchmark of cattle facial landmark detection on our CasttleFace-RGBT dataset using different backbones. The highest scores are shown in \textbf{bold}.}\label{tab:benchmark}
\resizebox{\linewidth}{!}{
\begin{NiceTabular}{c|l|ccc|*{3}{c}} 
\hline
  & \multirow{2}{*}{\textbf{Backbones}} &  \multicolumn{3}{c}{\textbf{Bounding Box}} & \multicolumn{3}{c}{\textbf{Keypoints}}\\
  \cmidrule(lr){3-5}\cmidrule(lr){6-8}
  & &  AP & $\text{AP}_{50}$ & $\text{AP}_{75}$ & AP & $\text{AP}_{50}$ & $\text{AP}_{75}$ \\
\hline
 \multirow{4}{*}{\rotatebox[]{90}{RGB}} & ResNet50 \cite{he2016deep} & 76.07 & 98.96 & 98.57 & 92.16 & \textbf{100.00} & 98.95 \\
 & ResNet101 \cite{he2016deep} & \textbf{76.12} & 98.94 & 98.54 & \textbf{94.37} & \textbf{100.00} & 98.94\\
 & ViT-B \cite{vit} & 74.33 & 98.96 & 98.87 & 93.85 & \textbf{100.00} & 98.93\\
 & Swin-B \cite{swin} & 75.28 & \textbf{98.99} & \textbf{98.97} & 93.21 & \textbf{100.00} & \textbf{100.00}\\
\hline \hline
\multirow{4}{*}{\rotatebox[]{90}{Thermal}} & ResNet50 \cite{he2016deep} &  65.18 & 99.45 & 77.39 & 53.07 & 97.89 & 39.50\\
 & ResNet101 \cite{he2016deep} & \textbf{72.30} & \textbf{100.00} & \textbf{98.88} & 64.60 & 98.02 & 67.71\\
 & ViT-B \cite{vit} & 55.41 & 98.99 & 50.73 & 46.92 & \textbf{99.98} & 31.83\\
 & Swin-B \cite{swin} & 66.81 & \textbf{100.00} & 81.99 & \textbf{73.16} & 99.01 & \textbf{79.75}\\
 \hline
\end{NiceTabular}}
\end{table}

\subsection{The Annotation Tool} 
\label{sec:annotation_tool}
While there are some off-the-shelf annotation tools on Github, we find them to be limited in helping us labeling. For example, DarkLabel cannot label individual points and it is closed source. Therefore, we create an annotation tool using C\# to help with the initial annotation and later correction process. 
The tool's design allows for the precise adjustment of each landmark according to the specified locations, ensuring accurate placement. If a point is at the wrong location, the annotator can click and drag that point to the correct position,
The annotated points are then recorded in a JSON file format, which includes the name of each image and a corresponding list of coordinate points for the facial keypoints.
As shown in Figure \ref{fig:tool}, the tool displays the image with landmarks overlayed on top.
The ``Browse" button at the top allows the user to specify what image will be used for annotation.
The  ``Next" and ``Prev" buttons allow the user to move through the specified images for correction.
The ``Save" button allows for the overwriting of the JSON file to store the annotated landmarks.
Finally, the "Discard" button marks unwanted images for removal after saving. 


\subsection{Landmarks} 
\label{sec:landmark}
The selection of 13 facial landmarks is driven by two primary motivations. Firstly, we aim to identify critical points for accurate temperature readings. To achieve this, we have chosen commonly recognized facial features such as the eyes, ears, muzzle, nostrils, and mouth. Secondly, we seek to capture the complete shape of the cow's face. To accomplish this, we have included additional landmarks specifically for the ears. These extra points ensure that the dataset encompasses the entire facial structure of the cattle, enabling a more comprehensive analysis of their facial morphology. Figure \ref{fig:face_part} illustrates the precise locations of the selected landmarks on the cattle's face.

\section{Benchmarking CattleFace-RGBT Dataset}

We perform the evaluation on bounding box and facial keypoint detection Feature Pyramid Network (FPN) \cite{fpn} architecture with different backbones with both CNNs-based networks (ResNet50 \cite{he2016deep} and ResNet101 \cite{he2016deep}) and
Transformer-based networks (ViT-B \cite{vit} and Swin-B \cite{swin}) in our CattleFace-RGBT dataset. We use Detectron2 framework~\cite{detectron2} to implement the training pipeline with a batch size of 16, AdamW \cite{adamw} optimizer with the learning rate is $1e-6$.
For both RGB and thermal, we randomly split the dataset into 70\% for traning and 30\% for testing. Following \cite{coco}, we use the standard average precision AP, $\text{AP}_{50}$, and $\text{AP}_{75}$, to evaluate the keypoint detection and bounding box detectrion performance. 

The results are shown in Table \ref{tab:benchmark}. For RGB images, it can be seen that ResNet101 achieve the highest bounding box and keypoint detection with 76.12 and 94.37 AP, respectively while Swin-B scores the highest $\text{AP}_{50}$ and $\text{AP}_{75}$ in bounding box detection. All models attain 100.00 $\text{AP}_{50}$ in keypoint detection, suggesting they perform exceptionally well at the 50\% IoU threshold. For thermal images, the results are generally lower, which could be due to the contrast in thermal images compared to RGB images. ResNet101 still obtains the highest scores across all evaluation metrics in bounding box detection, which indicates it is the most effective model for our dataset. 
In conclusion, ResNet101 and Swin-B are strong enough to be the baseline on our dataset.

\section{Conclusion}

In summary, we introduce the first dataset, called CattleFace-RGBT, containing both RGB and thermal images for cattle. The dataset includes 2,300 RGB and thermal image pairs, totaling 4,600 images. We provide annotations for 13 key points on key cattle facial parts in both the RGB and thermal images, including the eyes, ears, muzzle, nostrils, and mouth. Subsequently, we benchmark various keypoint detection methods on our dataset to demonstrate the performance of existing approaches. This will help researchers examine and choose appropriate baselines for future research and analysis on this novel cattle imagery dataset.
By releasing this novel dataset, we aim to advance research in the field of cattle welfare.

{
    \small
    \bibliographystyle{ieeenat_fullname}
    \bibliography{main}

\begin{thebibliography}{20}
\providecommand{\natexlab}[1]{#1}
\providecommand{\url}[1]{\texttt{#1}}
\expandafter\ifx\csname urlstyle\endcsname\relax
  \providecommand{\doi}[1]{doi: #1}\else
  \providecommand{\doi}{doi: \begingroup \urlstyle{rm}\Url}\fi

\bibitem[Cao et~al.(2019)Cao, Tang, Fang, Shen, Lu, and Tai]{jinkun19cross}
Jinkun Cao, Hongyang Tang, Hao-Shu Fang, Xiaoyong Shen, Cewu Lu, and Yu-Wing Tai.
\newblock {Cross-Domain Adaptation for Animal Pose Estimation}.
\newblock In \emph{ICCV}, 2019.

\bibitem[Deng et~al.(2019)Deng, Roussos, Chrysos, Ververas, Kotsia, Shen, and Zafeiriou]{deng2019menpo}
Jiankang Deng, Anastasios Roussos, Grigorios Chrysos, Evangelos Ververas, Irene Kotsia, Jie Shen, and Stefanos Zafeiriou.
\newblock The menpo benchmark for multi-pose 2d and 3d facial landmark localisation and tracking.
\newblock \emph{International Journal of Computer Vision}, 127, 2019.

\bibitem[Dosovitskiy et~al.(2021)Dosovitskiy, Beyer, Kolesnikov, Weissenborn, Zhai, Unterthiner, Dehghani, Minderer, Heigold, Gelly, Uszkoreit, and Houlsby]{vit}
Alexey Dosovitskiy, Lucas Beyer, Alexander Kolesnikov, Dirk Weissenborn, Xiaohua Zhai, Thomas Unterthiner, Mostafa Dehghani, Matthias Minderer, Georg Heigold, Sylvain Gelly, Jakob Uszkoreit, and Neil Houlsby.
\newblock {An Image is Worth 16x16 Words: Transformers for Image Recognition at Scale}.
\newblock In \emph{ICLR}, 2021.

\bibitem[Everingham et~al.()Everingham, Van~Gool, Williams, Winn, and Zisserman]{pascal-voc-2011}
M. Everingham, L. Van~Gool, C.~K.~I. Williams, J. Winn, and A. Zisserman.
\newblock The {PASCAL} {V}isual {O}bject {C}lasses {C}hallenge 2011 {(VOC2011)} {R}esults.
\newblock http://www.pascal-network.org/challenges/VOC/voc2011/workshop/index.html.

\bibitem[He et~al.(2016)He, Zhang, Ren, and Sun]{he2016deep}
Kaiming He, Xiangyu Zhang, Shaoqing Ren, and Jian Sun.
\newblock {Deep Residual Learning for Image Recognition}.
\newblock In \emph{CVPR}, 2016.

\bibitem[Khan et~al.(2020)Khan, McDonagh, Khan, Shahabuddin, Arora, Khan, Shao, and Tzimiropoulos]{animalweb}
Muhammad~Haris Khan, John McDonagh, Salman Khan, Muhammad Shahabuddin, Aditya Arora, Fahad~Shahbaz Khan, Ling Shao, and Georgios Tzimiropoulos.
\newblock {AnimalWeb: A Large-Scale Hierarchical Dataset of Annotated Animal Faces}.
\newblock In \emph{CVPR}, 2020.

\bibitem[Köstinger et~al.(2011)Köstinger, Wohlhart, Roth, and Bischof]{6130513}
Martin Köstinger, Paul Wohlhart, Peter~M. Roth, and Horst Bischof.
\newblock {Annotated Facial Landmarks in the Wild: A large-scale, real-world database for facial landmark localization}.
\newblock In \emph{IEEE International Conference on Computer Vision Workshops}, 2011.

\bibitem[Li et~al.(2023)Li, Ko, and Lee]{li2023feasibility}
Yuchuan Li, Yoon Ko, and Wonsook Lee.
\newblock A feasibility study on translation of rgb images to thermal images: Development of a machine learning algorithm.
\newblock \emph{SN Computer Science}, 4\penalty0 (5):\penalty0 555, 2023.

\bibitem[Lin et~al.(2014)Lin, Maire, Belongie, Hays, Perona, Ramanan, Doll{\'a}r, and Zitnick]{coco}
Tsung-Yi Lin, Michael Maire, Serge Belongie, James Hays, Pietro Perona, Deva Ramanan, Piotr Doll{\'a}r, and C~Lawrence Zitnick.
\newblock Microsoft coco: Common objects in context.
\newblock In \emph{ECCV}, 2014.

\bibitem[Lin et~al.(2017)Lin, Doll{\'a}r, Girshick, He, Hariharan, and Belongie]{fpn}
Tsung-Yi Lin, Piotr Doll{\'a}r, Ross Girshick, Kaiming He, Bharath Hariharan, and Serge Belongie.
\newblock Feature pyramid networks for object detection.
\newblock In \emph{CVPR}, 2017.

\bibitem[Liu et~al.(2012)Liu, Kanazawa, Jacobs, and Belhumeur]{liu2012dog}
Jiongxin Liu, Angjoo Kanazawa, David Jacobs, and Peter Belhumeur.
\newblock {Dog Breed Classification Using Part Localization}.
\newblock In \emph{ECCV}, 2012.

\bibitem[Liu et~al.(2021)Liu, Lin, Cao, Hu, Wei, Zhang, Lin, and Guo]{swin}
Ze Liu, Yutong Lin, Yue Cao, Han Hu, Yixuan Wei, Zheng Zhang, Stephen Lin, and Baining Guo.
\newblock {Swin Transformer: Hierarchical Vision Transformer using Shifted Windows}.
\newblock In \emph{ICCV}, 2021.

\bibitem[Loshchilov and Hutter(2017)]{adamw}
Ilya Loshchilov and Frank Hutter.
\newblock Decoupled weight decay regularization.
\newblock \emph{arXiv preprint arXiv:1711.05101}, 2017.

\bibitem[Martvel et~al.(2023)Martvel, Farhat, Shimshoni, and Zamansky]{catflw}
George Martvel, Nareed Farhat, Ilan Shimshoni, and Anna Zamansky.
\newblock {CatFLW: Cat Facial Landmarks in the Wild Dataset}.
\newblock \emph{arXiv preprint arXiv:2305.04232}, 2023.

\bibitem[Ong et~al.(2023)Ong, Retta, Srinivasan, Tan, and Liu]{cattleeye}
Kian~Eng Ong, Sivaji Retta, Ramarajulu Srinivasan, Shawn Tan, and Jun Liu.
\newblock {CattleEyeView: A Multi-task Top-down View Cattle Dataset for Smarter Precision Livestock Farming}.
\newblock In \emph{VCIP}, 2023.

\bibitem[Rashid et~al.(2017)Rashid, Gu, and Lee]{horse}
Maheen Rashid, Xiuye Gu, and Yong~Jae Lee.
\newblock {Interspecies Knowledge Transfer for Facial Keypoint Detection}.
\newblock In \emph{CVPR}, 2017.

\bibitem[Sagonas et~al.(2013)Sagonas, Tzimiropoulos, Zafeiriou, and Pantic]{6755925}
Christos Sagonas, Georgios Tzimiropoulos, Stefanos Zafeiriou, and Maja Pantic.
\newblock {300 Faces in-the-Wild Challenge: The First Facial Landmark Localization Challenge}.
\newblock In \emph{IEEE International Conference on Computer Vision Workshops}, 2013.

\bibitem[Wu et~al.(2019)Wu, Kirillov, Massa, Lo, and Girshick]{detectron2}
Yuxin Wu, Alexander Kirillov, Francisco Massa, Wan-Yen Lo, and Ross Girshick.
\newblock Detectron2.
\newblock \url{https://github.com/facebookresearch/detectron2}, 2019.

\bibitem[Yang et~al.(2016)Yang, Zhang, and Robinson]{yang15sheep}
Heng Yang, Renqiao Zhang, and Peter Robinson.
\newblock {Human and Sheep Facial Landmarks Localisation by Triplet Interpolated Features}.
\newblock In \emph{WACV}, 2016.

\bibitem[Zhang et~al.(2008)Zhang, Sun, and Tang]{cathead}
Weiwei Zhang, Jian Sun, and Xiaoou Tang.
\newblock Cat head detection - how to effectively exploit shape and texture features.
\newblock \emph{ECCV}, 2008.

\end{thebibliography}
}


\end{document}